\documentclass[11pt,a4paper]{article}
\usepackage[hyperref]{emnlp2018}
\usepackage{times}
\usepackage{latexsym}
\usepackage{graphicx}

\usepackage{url}

\usepackage{bm}
\usepackage{amsfonts}

\def\argmax{\mathop{\rm argmax}}%

\aclfinalcopy % Uncomment this line for the final submission
\newcommand\redbf[1]{\textcolor{red}{\textbf{#1}}}
\newcommand\bluebf[1]{\textcolor{blue}{\textbf{#1}}}

\title{Improving the Transformer Translation Model \\ with Document-Level Context}

\author{Jiacheng Zhang$^\dagger$,  Huanbo Luan$^\dagger$, Maosong Sun$^{\dagger}$, FeiFei Zhai$^\#$,\\ \bf{Jingfang Xu$^\#$, Min Zhang$^\S$ and Yang Liu}$^{\dagger\ddagger}$\thanks{\ \ Corresponding author: Yang Liu.} \\
    $^\dagger$Institute for Artificial Intelligence \\
    State Key Laboratory of Intelligent Technology and Systems \\
    Department of Computer Science and Technology, Tsinghua University, Beijing, China \\
    $^\ddagger$Beijing National Research Center for Information Science and Technology\\
    $^\#$Sogou Inc., Beijing, China\\
    $^\S$Soochow University, Suzhou, China\\
    }

\date{}
\begin{document}
\maketitle
\begin{abstract}
Although the Transformer translation model \cite{Vaswani:17} has achieved state-of-the-art performance in a variety of translation tasks, how to use document-level context to deal with discourse phenomena problematic for Transformer still remains a challenge. In this work, we extend the Transformer model with a new context encoder to represent document-level context, which is then incorporated into the original encoder and decoder. As large-scale document-level parallel corpora are usually not available, we introduce a two-step training method to take full advantage of abundant sentence-level parallel corpora and limited document-level parallel corpora. Experiments on the NIST Chinese-English datasets and the IWSLT French-English datasets show that our approach improves over Transformer significantly. \footnote{The source code is available at \url{https://github.com/Glaceon31/Document-Transformer}}
\end{abstract}

\section{Introduction}

The past several years have witnessed the rapid development of neural machine translation (NMT) \cite{Sutskever:14,Bahdanau:15}, which investigates the use of neural networks to model the translation process. Showing remarkable superiority over conventional statistical machine translation (SMT), NMT has been recognized as the new {\em de facto} method and is widely used in commercial MT systems \cite{Wu:16}. A variety of NMT models have been proposed to map between natural languages such as RNNencdec \cite{Sutskever:14}, RNNsearch \cite{Bahdanau:15}, ConvS2S \cite{Gehring:17}, and Transformer \cite{Vaswani:17}. Among them, the Transformer model has achieved state-of-the-art translation performance. The capability to minimize the path length between long-distance dependencies in neural networks contributes to its exceptional performance.

However, the Transformer model still suffers from a major drawback: it performs translation only at the sentence level and ignores document-level context. Document-level context has proven to be beneficial for improving translation performance, not only for conventional SMT \cite{Gong:11,Hardmeier:12}, but also for NMT \cite{Wang:17,Tu:18}. \citet{Bawden:18} indicate that it is important to exploit document-level context to deal with context-dependent phenomena which are problematic for machine translation such as coreference, lexical cohesion, and lexical disambiguation.

While document-level NMT has attracted increasing attention from the community in the past two years \cite{Jean:17,Kuang:17,Tiedemann:17,Wang:17,Maruf:18,Bawden:18,Tu:18,Voita:18}, to the best of our knowledge, only one existing work has endeavored to model document-level context for the Transformer model \cite{Voita:18}. Previous approaches to document-level NMT have concentrated on the RNNsearch model \cite{Bahdanau:15}. It is challenging to adapt these approaches to Transformer because they are designed specifically for RNNsearch.

\begin{figure*}[!t]
    \centering
    \includegraphics[width=0.9\textwidth]{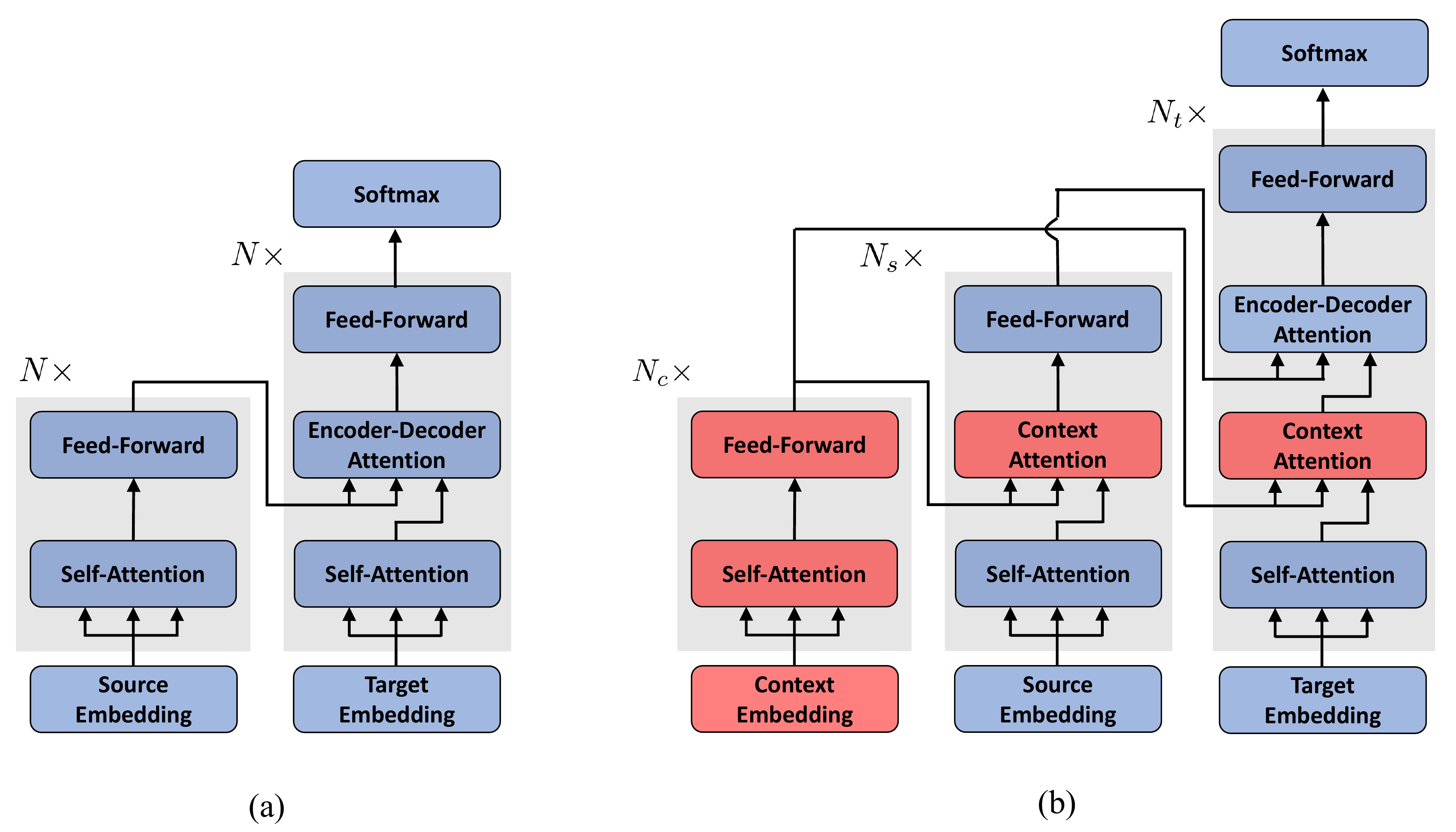}
    \caption{(a) The original Transformer translation model \cite{Vaswani:17} and (b) the extended Transformer translation model that exploits document-level context. The newly introduced modules are highlighted in red.}
    \label{fig1}
\end{figure*}

In this work, we propose to extend the Transformer model to take advantage of document-level context. The basic idea is to use multi-head self-attention \cite{Vaswani:17} to compute the representation of document-level context, which is then incorporated into the encoder and decoder using multi-head attention. Since large-scale document-level parallel corpora are usually hard to acquire, we propose to train sentence-level model parameters on sentence-level parallel corpora first and then estimate document-level model parameters on document-level parallel corpora while keeping the learned original sentence-level Transformer model parameters fixed. Our approach has the following advantages:

\begin{enumerate}
\item {\em Increased capability to capture context}: the use of multi-head attention, which significantly reduces the path length between long-range dependencies, helps to improve the capability to capture document-level context; 
\item {\em Small computational overhead}: as all newly introduced modules are based on highly parallelizable multi-head attention, there is no significant slowdown in both training and decoding;
\item {\em Better use of limited labeled data}: our approach is capable of maintaining the superiority over the sentence-level counterpart even when only small-scale document-level parallel corpora are available. 
\end{enumerate}

Experiments show that our approach achieves an improvement of 1.96 and 0.89 BLEU points over Transformer on Chinese-English and French-English translation respectively by exploiting document-level context. It also outperforms a state-of-the-art cache-based method \cite{Kuang:17} adapted for Transformer.

\section{Approach}

\subsection{Problem Statement}

Our goal is to enable the Transformer translation model \cite{Vaswani:17} as shown in Figure \ref{fig1}(a) to exploit document-level context.

Formally, let $\mathbf{X} = \mathbf{x}^{(1)}, \dots, \mathbf{x}^{(k)}, \dots, \mathbf{x}^{(K)}$ be a source-language document composed of $K$ source sentences. We use $\mathbf{x}^{(k)} = x^{(k)}_1, \dots, x^{(k)}_i, \dots, x^{(k)}_I$ to denote the $k$-th source sentence containing $I$ words. $x^{(k)}_i$ denotes the $i$-th word in the $k$-th source sentence.  Likewise, the corresponding target-language document is denoted by $\mathbf{Y} = \mathbf{y}^{(1)}, \dots, \mathbf{y}^{(k)}, \dots, \mathbf{y}^{(K)}$ and $\mathbf{y}^{(k)}=y^{(k)}_1,\dots, y^{(k)}_j, \dots, y^{(k)}_J$ represents the $k$-th target sentence containing $J$ words. $y^{(k)}_j$ denotes the $j$-th word in the $k$-th target sentence. We assume that $\langle \mathbf{X}, \mathbf{Y} \rangle$ constitutes a {\em parallel document} and each $\langle \mathbf{x}^{(k)}, \mathbf{y}^{(k)} \rangle$ forms a {\em parallel sentence}.

Therefore, the document-level translation probability is given by
\begin{eqnarray}
P(\mathbf{Y}|\mathbf{X}; \bm{\theta}) = \prod_{k=1}^{K} P(\mathbf{y}^{(k)}|\mathbf{X}, \mathbf{Y}_{<k}; \bm{\theta}), \label{eq:original_document_prob}
\end{eqnarray}
where $\mathbf{Y}_{<k} = \mathbf{y}^{(1)}, \dots, \mathbf{y}^{(k-1)}$ is a partial translation.

For generating $\mathbf{y}^{(k)}$, the source document $\mathbf{X}$ can be divided into three parts: (1) the $k$-th source sentence $\mathbf{X}_{=k} = \mathbf{x}^{(k)}$, (2) the source-side document-level context on the left $\mathbf{X}_{<k} =\mathbf{x}^{(1)},\dots, \mathbf{x}^{(k-1)}$, and (3) the source-side document-level context on the right $\mathbf{X}_{>k} = \mathbf{x}^{(k+1)}, \dots, \mathbf{x}^{(K)}$. As the languages used in our experiments (i.e., Chinese and English) are written left to right, we omit $\mathbf{X}_{>k}$ for simplicity. 

We also omit the target-side document-level context $\mathbf{Y}_{<k}$ due to the translation error propagation problem \cite{Wang:17}: errors made in translating one sentence will be propagated to the translation process of subsequent sentences. Interestingly, we find that using source-side document-level context $\mathbf{X}_{<k}$, which conveys the same information with $\mathbf{Y}_{<k}$, helps to compute better representations on the target side (see Table \ref{table:integration}).

As a result, the document-level translation probability can be approximated as
\begin{eqnarray}
&& P(\mathbf{Y}|\mathbf{X}; \bm{\theta}) \nonumber \\
&\approx& \prod_{k=1}^{K} P(\mathbf{y}^{(k)}|\mathbf{X}_{<k}, \mathbf{x}^{(k)}; \bm{\theta}), \\
&=& \prod_{k=1}^{K} \prod_{j=1}^{J} P(y^{(k)}_j | \mathbf{X}_{<k}, \mathbf{x}^{(k)}, \mathbf{y}^{(k)}_{<j}; \bm{\theta}),
\end{eqnarray}
where $\mathbf{y}^{(k)}_{<j} = y^{(k)}_1, \dots, y^{(k)}_{j-1}$ is a partial translation.

In this way, the document-level translation model can still be defined at the sentence level without sacrificing efficiency except that the source-side document-level context $\mathbf{X}_{<k}$ (or {\em context} for short) is  taken into account.

In the following, we will introduce how to represent the context (Section \ref{sec:representation}), how to integrate the context (Section \ref{sec:integration}), and how to train the model especially when only limited training data is available (Section \ref{sec:training}).

\subsection{Document-level Context Representation} \label{sec:representation}

As document-level context often includes several sentences, it is important to capture long-range dependencies and identify relevant information. We use multi-head self-attention \cite{Vaswani:17} to compute the representation of document-level context because it is capable of reducing the maximum path length between long-range dependencies to $O(1)$ \cite{Vaswani:17} and determining the relative importance of different locations in the context \cite{Bahdanau:15}. Because of this property, multi-head self-attention has proven to be effective in other NLP tasks such as constituency parsing \cite{Kitaev:18}.

As shown in Figure \ref{fig1}(b), we use a self-attentive encoder to compute the representation of $\mathbf{X}_{<k}$. The input to the self-attentive encoder is a sequence of context word embeddings, represented as a matrix. Suppose $\mathbf{X}_{<k}$ is composed of $M$ source words: $\mathbf{X}_{<k} = x_1, \dots, x_m, \dots, x_M$. We use $\bm{x}_m \in \mathbb{R}^{D \times 1}$ to denote the vector representation of $x_m$ that is the sum of word embedding and positional encoding \cite{Vaswani:17}. Therefore, the matrix representation of $\mathbf{X}_{<k}$ is given by 
\begin{eqnarray}
\bm{X}_c = [\bm{x}_1; \dots; \bm{x}_M],
\end{eqnarray} 
where $\bm{X}_c \in \mathbb{R}^{D \times M}$ is the concatenation of all vector representations of all source contextual words.

The self-attentive encoder is composed of a stack of $N_c$ identical layers. Each layer has two sub-layers. The first sub-layer is a multi-head self-attention:
\begin{eqnarray}
\mathbf{A}^{(1)} = \mathrm{MultiHead}(\bm{X}_c, \bm{X}_c, \bm{X}_c),
\end{eqnarray}
where $\mathbf{A}^{(1)} \in \mathbb{R}^{D \times M}$ is the hidden state calculated by the multi-head self-attention at the first layer, $\mathrm{MultiHead}(\mathbf{Q}, \mathbf{K}, \mathbf{V})$ is  a multi-head self-attention function that takes a query matrix $\mathbf{Q}$, a key matrix $\mathbf{K}$, and a value matrix $\mathbf{V}$ as inputs. In this case, $\mathbf{Q} = \mathbf{K} = \mathbf{V} = \bm{X}_c$. This is why it is called self-attention. Please refer to \cite{Vaswani:17} for more details. 

Note that we follow \citet{Vaswani:17} to use residual connection and layer normalization in each sub-layer, which are omitted in the presentation for simplicity. For example, the actual output of the first sub-layer is:
\begin{eqnarray}
\mathrm{LayerNorm}(\mathbf{A}^{(1)}+\bm{X}_c).
\end{eqnarray}

The second sub-layer is a simple, position-wise fully connected feed-forward network:
\begin{eqnarray}
\mathbf{C}^{(1)} = \Big[ \mathrm{FNN}(\mathbf{A}^{(1)}_{\cdot, 1}); \dots; \mathrm{FNN}(\mathbf{A}^{(1)}_{\cdot, M}) \Big]
\end{eqnarray}
where $\mathbf{C}^{(1)} \in \mathbb{R}^{D \times M}$ is the annotation of $\mathbf{X}_{<k}$ after the first layer, $\mathbf{A}^{(1)}_{\cdot, m} \in \mathbb{R}^{D \times 1}$ is the column vector for the $m$-th contextual word, and $\mathrm{FNN}(\cdot)$ is a position-wise fully connected feed-forward network \cite{Vaswani:17}.

This process iterates $N_c$ times as follows:
\begin{eqnarray}
\mathbf{A}^{(n)} = \mathrm{MultiHead}\Big(\mathbf{C}^{(n-1)}, \mathbf{C}^{(n-1)}, \mathbf{C}^{(n-1)}\Big), \\
\mathbf{C}^{(n)} = \Big[ \mathrm{FNN}(\mathbf{A}^{(n)}_{\cdot, 1}); \dots; \mathrm{FNN}(\mathbf{A}^{(n)}_{\cdot, M}) \Big], \quad \quad \ \
\end{eqnarray}
where $\mathbf{A}^{(n)}$ and $\mathbf{C}^{(n)}$ ($n=1,\dots, N_c$) are the hidden state and annotation at the $n$-th layer, respectively. Note that $\mathbf{C}^{(0)} = \bm{X}_c$.

\subsection{Document-level Context Integration}\label{sec:integration}

We use multi-head attention to integrate $\mathbf{C}^{(N_c)}$, which is the representation of $\mathbf{X}_{<k}$, into both the encoder and the decoder.

\subsubsection{Integration into the Encoder}
Given the $k$-th source sentence $\mathbf{x}^{(k)}$, we use $\bm{x}^{(k)}_i \in \mathbb{R}^{D \times 1}$ to  denote the vector representation of the $i$-th source word $x^{(k)}_i$, which is a sum of word embedding and positional encoding. Therefore, the initial matrix representation of $\mathbf{x}^{(k)}$ is
\begin{eqnarray}
\bm{X} = [\bm{x}^{(k)}_1;\dots; \bm{x}^{(k)}_I],
\end{eqnarray}
where $\bm{X} \in \mathbb{R}^{D \times I}$ is the concatenation of all vector representations of source words.

As shown in Figure \ref{fig1}(b), we follow \cite{Vaswani:17} to use a stack of $N_s$ identical layers to encode $\mathbf{x}^{(k)}$. Each layer consists of three sub-layers. The first sub-layer is a multi-head self-attention:
\begin{eqnarray}
\mathbf{B}^{(n)} = \mathrm{MultiHead}\Big(\mathbf{S}^{(n-1)}, \mathbf{S}^{(n-1)}, \mathbf{S}^{(n-1)} \Big),
\end{eqnarray}
where $\mathbf{S}^{(0)} = \bm{X}$.
The second sub-layer is context attention that integrates document-level context into the encoder:
\begin{eqnarray}
\mathbf{D}^{(n)} = \mathrm{MultiHead}\Big(\mathbf{B}^{(n)}, \mathbf{C}^{(N_c)}, \mathbf{C}^{(N_c)} \Big). \label{eq:ctx_enc}
\end{eqnarray}
The third sub-layer is a position-wise fully connected feed-forward neural network:
\begin{eqnarray}
\mathbf{S}^{(n)} = \Big[ \mathrm{FNN}(\mathbf{D}^{(n)}_{\cdot, 1}); \dots; \mathrm{FNN}(\mathbf{D}^{(n)}_{\cdot, I}) \Big], 
\end{eqnarray}
where $\mathbf{S}^{(n)} \in \mathbb{R}^{D \times I}$ is the representation of the source sentence $\mathbf{x}^{(k)}$ at the $n$-th layer ($n=1,\dots, N_s$).

\subsubsection{Integration into the Decoder}
When generating the $j$-th target word $y^{(k)}_j$, the partial translation is denoted by $\mathbf{y}^{(k)}_{<j} = y^{(k)}_1,\dots,y^{(k)}_{j-1}$. We follow \citet{Vaswani:17} to offset the target word embeddings by one position, resulting in the following matrix representation of $\mathbf{y}^{(k)}_{<j}$:
\begin{eqnarray}
\bm{Y} = \big[ \bm{y}^{(k)}_0, \dots, \bm{y}^{(k)}_{j-1} \big],
\end{eqnarray}
where $\mathbf{y}^{(k)}_0 \in \mathbb{R}^{D \times 1}$ is the vector representation of a begin-of-sentence token and $\bm{Y} \in \mathbb{R}^{D \times j}$ is the concatenation of all vectors.

As shown in Figure \ref{fig1}(b),  we follow \cite{Vaswani:17} to use a stack of $N_t$ identical layers to compute target-side representations. Each layer is composed of four sub-layers. The first sub-layer is a multi-head self-attention:
\begin{eqnarray}
\mathbf{E}^{(n)} = \mathrm{MultiHead}\Big(\mathbf{T}^{(n-1)}, \mathbf{T}^{(n-1)}, \mathbf{T}^{(n-1)}\Big), 
\end{eqnarray}
where $\mathbf{T}^{(0)} = \bm{Y}$.
The second sub-layer is context attention that integrates document-level context into the decoder:
\begin{eqnarray}
\mathbf{F}^{(n)} = \mathrm{MultiHead}\Big(\mathbf{E}^{(n)}, \mathbf{C}^{(N_c)}, \mathbf{C}^{(N_c)}\Big). \label{eq:ctx_dec}
\end{eqnarray}
The third sub-layer is encoder-decoder attention that integrates the representation of the corresponding source sentence:
\begin{eqnarray}
\mathbf{G}^{(n)} = \mathrm{MultiHead}\Big(\mathbf{F}^{(n)}, \mathbf{S}^{(N_s)}, \mathbf{S}^{(N_s)}\Big).
\end{eqnarray}
The fourth sub-layer is a position-wise fully connected feed-forward neural network:
\begin{eqnarray}
\mathbf{T}^{(n)} = \Big[ \mathrm{FNN}(\mathbf{G}^{(n)}_{\cdot, 1}); \dots; \mathrm{FNN}(\mathbf{G}^{(n)}_{\cdot, j}), \Big],
\end{eqnarray}
where $\mathbf{T}^{(n)} \in \mathbb{R}^{D \times j}$ is the representation at the $n$-th layer ($n=1,\dots, N_t$). Note that $\mathbf{T}^{(0)} = \bm{Y}$.

Finally, the probability distribution of generating the next target word $y^{(k)}_j$ is defined using a softmax layer:
\begin{eqnarray}
P(y^{(k)}_j | \mathbf{X}_{<k}, \mathbf{x}^{(k)}, \mathbf{y}^{(k)}_{<j}; \bm{\theta}) \propto \exp(\mathbf{W}_{o} \mathbf{T}^{(N_t)}_{\cdot, j})
\end{eqnarray}
where $\mathbf{W}_{o} \in \mathbb{R}^{|\mathcal{V}_y| \times D}$ is a model parameter, $\mathcal{V}_y$ is the target vocabulary, and $\mathbf{T}^{(N_t)}_{\cdot, j} \in \mathbb{R}^{D \times 1}$ is a column vector for predicting the $j$-th target word.

\subsubsection{Context Gating} \label{sec:gating}
In our model, we follow \citet{Vaswani:17} to use residual connections \cite{He:16} around each sub-layer to shortcut its input to its output:
\begin{eqnarray}
\mathrm{Residual}(\mathbf{H}) = \mathbf{H} + \mathrm{SubLayer}(\mathbf{H}),
\end{eqnarray}
where $\mathbf{H}$ is the input of the sub-layer.

While residual connections prove to be effective for building deep architectures, there is one potential problem for our model: the residual connections after the context attention sub-layer might increase the influence of document-level context $\mathbf{X}_{<k}$ in an uncontrolled way. This is undesirable because the source sentence $\mathbf{x}^{(k)}$ usually plays a more important role in target word generation.

To address this problem, we replace the residual connections after the context attention sub-layer with a {\em position-wise context gating} sub-layer:
\begin{eqnarray}
\mathrm{Gating}(\mathbf{H}) = \lambda \mathbf{H} + (1 - \lambda) \mathrm{SubLayer}(\mathbf{H}).
\end{eqnarray}
The gating weight is given by
\begin{eqnarray}
\lambda = \sigma(\mathbf{W}_i \mathbf{H} + \mathbf{W}_s \mathrm{SubLayer}(\mathbf{H})),
\end{eqnarray}
where $\sigma(\cdot)$ is a sigmoid function, $\mathbf{W}_i$ and $\mathbf{W}_s$ are model parameters.

\subsection{Training}
\label{sec:training}

Given a document-level parallel corpus $D_d$, the standard training objective is to maximize the log-likelihood of the training data:
\begin{eqnarray}
\hat{\bm{\theta}} = \argmax_{\bm{\theta}} \Bigg\{ \sum_{\langle \mathbf{X}, \mathbf{Y} \rangle \in D_d} \log P(\mathbf{Y}|\mathbf{X}; \bm{\theta}) \Bigg\}. \label{eq:direct}
\end{eqnarray}

Unfortunately, large-scale document-level parallel corpora are usually unavailable, even for resource-rich languages such as English and Chinese. Under small-data training conditions, document-level NMT is prone to underperform sentence-level NMT because of poor estimates of low-frequency events.

To address this problem, we adopt the idea of freezing some parameters while tuning the remaining part of the model \cite{Jean:15,Zoph:16}. We propose a two-step training strategy that uses an additional sentence-level parallel corpus $D_s$, which can be larger than $D_d$. We divide model parameters into two subsets: $\bm{\theta} = \ \bm{\theta}_s \cup \bm{\theta}_d $, where $\bm{\theta}_s$ is a set of original sentence-level model parameters (highlighted in blue in Figure \ref{fig1}(b)) and $\bm{\theta}_d$ is a set of newly-introduced document-level model parameters (highlighted in red in Figure \ref{fig1}(b)).

In the first step, sentence-level parameters $\bm{\theta}_s$ are estimated on the combined sentence-level parallel corpus $D_s \cup D_d$: \footnote{It is easy to create a sentence-level parallel corpus from $D_d$.}
\begin{eqnarray}
\hat{\bm{\theta}_s} = \argmax_{\bm{\theta}_s} \sum_{\langle \mathbf{x}, \mathbf{y} \rangle \in D_s \cup D_d} \log P(\mathbf{y}|\mathbf{x}; \bm{\theta}_s). \label{eq:sent}
\end{eqnarray}
Note that the newly introduced modules (highlighted in red in Figure \ref{fig1}(b)) are inactivated in this step. $P(\mathbf{y}|\mathbf{x}; \bm{\theta}_s)$ is identical to the original Transformer model, which is a special case of our model.

In the second step, document-level parameters $\bm{\theta}_d$ are estimated on the document-level parallel corpus $D_d$ only:
\begin{eqnarray}
\hat{\bm{\theta}_d} = \argmax_{\bm{\theta}_d} \sum_{\langle \mathbf{X}, \mathbf{Y} \rangle \in D_d} \log P(\mathbf{Y} | \mathbf{X}; \hat{\bm{\theta}}_s, \bm{\theta}_d). \label{eq:doc}
\end{eqnarray}

Our approach is also similar to pre-training which has been widely used in NMT \cite{Shen:16,Tu:18}. The major difference is that our approach keeps $\hat{\bm{\theta}}_{s}$ {\em fixed} when estimating $\bm{\theta}_d$ to prevent the model from overfitting on the relatively smaller document-level parallel corpora.

\section{Experiments}

\subsection{Setup}
We evaluate our approach on Chinese-English and French-English translation tasks. In Chinese-English translation task, the training set contains 2M Chinese-English sentence pairs with 54.8M Chinese words and 60.8M English words. \footnote{The training set consists of sentence-level parallel corpora LDC2002E18, LDC2003E07, LDC2003E14, news part of LDC2004T08 and document-level parallel corpora LDC2002T01, LDC2004T07, LDC2005T06, LDC2005T10, LDC2009T02, LDC2009T15, LDC2010T03.} The document-level parallel corpus is a subset of the full training set, including 41K documents with 940K sentence pairs. On average, each document in the training set contains 22.9 sentences. We use the NIST 2006 dataset as the development set and the NIST 2002, 2003, 2004, 2005, 2008 datasets as test sets. The development and test sets contain 588 documents with 5,833 sentences. On average, each document contains 9.9 sentences. 

In French-English translation task, we use the IWSLT bilingual training data \cite{Mauro:12} which contains 1,824 documents with 220K sentence pairs as training set. For development and testing, we use the IWSLT 2010 development and test sets, which contains 8 documents with 887 sentence pairs and 11 documents with 1,664 sentence pairs respectively. The evaluation metric for both tasks is case-insensitive BLEU score as calculated by the {\em multi-bleu.perl} script.

In preprocessing, we use byte pair encoding \cite{Sennrich:16} with 32K merges to segment words into sub-word units for all languages. For the original Transformer model and our extended model, the hidden size is set to 512 and the filter size is set to 2,048. The multi-head attention has 8 individual attention heads. We set $N=N_s=N_t=6$. In training, we use Adam \cite{Kingma:15} for optimization. Each mini-batch contains approximately 24K words. We use the learning rate decay policy described by \citet{Vaswani:17}. In decoding, the beam size is set to 4. We use the length penalty \cite{Wu:16} and set the hyper-parameter $\alpha$ to 0.6. We use four Tesla P40 GPUs for training and one Tesla P40 GPU for decoding. We implement our approach on top of the open-source toolkit THUMT \cite{Zhang:17}.  \footnote{\url{https://github.com/thumt/THUMT}}

\subsection{Effect of Context Length}

\begin{table}[!t]
\centering
\begin{tabular}{|c|c|c|c|}
\hline
\# sent. & 1 & 2 & 3 \\
\hline \hline
MT06 & 49.38 & \textbf{49.69} & 49.49 \\
\hline
\end{tabular}
\caption{Effect of context length on translation quality. The BLEU scores are calculated on the development set.} \label{table:effect_context_length}
\end{table}

We first investigate the effect of context length (i.e., the number of preceding sentences) on our approach. As shown in Table \ref{table:effect_context_length}, using two preceding source sentences as document-level context achieves the best translation performance on the development set. Using more preceding sentences does not bring any improvement and increases computational cost. This confirms the finding of \citet{Tu:18} that long-distance context only has limited influence. Therefore, we set the number of preceding sentences to 2 in the following experiments. \footnote{If there is no preceding sentence, we simply use a single begin-of-sentence token.}

\subsection{Effect of Self-Attention Layer Number}

\begin{table}[!t]
\centering
\begin{tabular}{|c|c|}
\hline
\# Layer & MT06 \\
\hline \hline
1 & \textbf{49.69} \\
2 & 49.38 \\
3 & 49.54 \\
4 & 49.59 \\
5 & 49.31 \\
6 & 49.43 \\
\hline
\end{tabular}
\caption{Effect of self-attention layer number (i.e., $N_c$) on translation quality. The BLEU scores are calculated on the development set.} \label{table:effect_layer_number}
\end{table}

Table \ref{table:effect_layer_number} shows the effect of self-attention layer number for computing representations of document-level context (see Section \ref{sec:representation}) on translation quality. Surprisingly, using only one self-attention layer suffices to achieve good performance. Increasing the number of self-attention layers does not lead to any improvements. Therefore, we set $N_c$ to 1 for efficiency.

\subsection{Comparison with Previous Work}

\begin{table*}[!th]
\centering
\begin{tabular}{|c|c||c|ccccc|c|}
\hline
Method & Model & MT06 & MT02 & MT03 & MT04 & MT05 & MT08 & All \\
\hline \hline
\cite{Wang:17} & RNNsearch &  37.76 & - & - & - & 36.89 & 27.57 & - \\
\cite{Kuang:17} & RNNsearch & - & 34.41 & - & 38.40 & 32.90 & 31.86 & - \\
\hline \hline
\cite{Vaswani:17} & Transformer & 48.09 & 48.63 & 47.54 & 47.79 & 48.34 & 38.31 & 45.97 \\
\cite{Kuang:17}* & Transformer  & 48.14 & 48.97 & 48.05 & 47.91 & 48.53 & 38.38 & 46.37 \\
\hline \hline
{\em this work} & Transformer  & \textbf{49.69} & \textbf{50.96} & \textbf{50.21} & \textbf{49.73} & \textbf{49.46} & \textbf{39.69} & \textbf{47.93} \\
\hline
\end{tabular}
\caption{Comparison with previous works on Chinese-English translation task. The evaluation metric is case-insensitive BLEU score. \cite{Wang:17} use a hierarchical RNN to incorporate document-level context into RNNsearch. \cite{Kuang:17} use a cache to exploit document-level context for RNNsearch. \cite{Kuang:17}* is an adapted version of the cache-based method for Transformer. Note that ``MT06'' is not included in ``All''.} \label{table:comparison}
\end{table*}

\begin{table}[!t]
\centering
\begin{tabular}{|c||c|c|}
\hline
Method & Dev & Test \\
\hline \hline
Transformer & 29.42 & 35.15 \\
{\em this work} & \textbf{30.40} & \textbf{36.04} \\
\hline
\end{tabular}
\caption{Comparison with Transformer on French-English translation task. The evaluation metric is case-insensitive BLEU score.} \label{table:fren}
\end{table}

In Chinese-English translation task, we compare our approach with the following previous methods:
\begin{enumerate}
\item \cite{Wang:17}: using a hierarchical RNN to integrate document-level context into the RNNsearch model. They use a document-level parallel corpus containing 1M sentence pairs. Table \ref{table:comparison} gives the BLEU scores reported in their paper.

\item \cite{Kuang:17}: using a cache which stores previous translated words and topical words to incorporate document-level context into the RNNsearch model. They use a document-level parallel corpus containing 2.8M sentence pairs. Table \ref{table:comparison} gives the BLEU scores reported in their paper.

\item \cite{Vaswani:17}: the state-of-the-art NMT model that does not exploit document-level context. We use the open-source toolkit THUMT \cite{Zhang:17} to train and evaluate the model. The training dataset is our sentence-level parallel corpus containing 2M sentence pairs.

\item \cite{Kuang:17}*: adapting the cache-based method to the Transformer model. We implement it on top of the open-source toolkit THUMT. We also use the same training data (i.e., 2M sentence pairs) and the same two-step training strategy to estimate sentence- and document-level parameters separately.

\end{enumerate}

As shown in Table \ref{table:comparison}, using the same data, our approach achieves significant improvements over the original Transformer model \cite{Vaswani:17} ($p < 0.01$). The gain on the concatenated test set (i.e., ``All'') is 1.96 BLEU points. It also outperforms the cache-based method \cite{Kuang:17} adapted for Transformer significantly ($p < 0.01$), which also uses the two-step training strategy.
Table \ref{table:fren} shows that our model also outperforms Transformer by 0.89 BLEU points on French-English translation task. 

\subsection{Subjective Evaluation}

\begin{table}[!t]
\centering
\begin{tabular}{|c||c|c|c|}
\hline
& $>$ & $=$ & $<$ \\
\hline \hline
Human 1 & 24\% & 45\% & 31\% \\
Human 2 & 20\% & 55\% & 25\% \\
Human 3 & 12\% & 52\% & 36\% \\
\hline \hline
Overall & 19\% & 51\% & 31\% \\
\hline
\end{tabular}
\caption{Subjective evaluation of the comparison between the original Transformer model and our model. ``$>$'' means that Transformer is better than our model, ``$=$'' means equal, and ``$<$'' means worse.} \label{table:subjective}
\end{table}

We also conducted a subjective evaluation to validate the benefit of exploiting document-level context. All three human evaluators were asked to compare the outputs of the original Transformer model and our model of 20 documents containing 198 sentences, which were randomly sampled from the test sets.

Table \ref{table:subjective} shows the results of subjective evaluation. Three human evaluators generally made consistent judgements. On average, around 19\% of Transformer's translations are better than that of our model, 51\% are equal, and 31\% are worse. This evaluation confirms that exploiting document-level context helps to improve translation quality.

\subsection{Evaluation of Efficiency}

\begin{table}[!t]
\centering
\begin{tabular}{|c||c|c|}
\hline
Method & Training & Decoding \\
\hline \hline
Transformer & 41K & 872\\
{\em this work} & 31K & 364\\
\hline
\end{tabular}
\caption{Evaluation of training and decoding speed. The speed is measured in terms of word/second (wps).} \label{table:efficiency}
\end{table}

We evaluated the efficiency of our approach. It takes the original Transformer model about 6.7 hours to converge during training and the training speed is 41K words/second. The decoding speed is 872 words/second. In contrast, it takes our model about 7.8 hours to converge in the second step of training. The training speed is 31K words/second. The decoding speed is 364 words/second.

Therefore, the training speed is only reduced by 25\% thanks to the high parallelism of multi-head attention used to incorporate document-level context. The gap is larger in decoding because target words are generated in an autoregressive way in Transformer.

\subsection{Effect of Two-Step Training}

\begin{table*}[!t]
\centering
\begin{tabular}{|cc||c|ccccc|c|}
\hline
sent. & doc. & MT06 & MT02 & MT03 & MT04 & MT05 & MT08 & All \\
\hline \hline
940K & - & 36.20  & 42.41 & 43.12 & 41.02 & 40.93 & 31.49 & 39.53 \\
2M & - & 48.09 & 48.63 & 47.54 & 47.79 & 48.34 & 38.31 & 45.97 \\
- & 940K & 34.00 & 38.83 & 40.51 & 38.30 & 36.69 & 29.38 & 36.52 \\
940K  & 940K & 37.12 & 43.29 & 43.70 & 41.42 & 41.84 & 32.36 & 40.22 \\
2M  & 940K & 49.69  & 50.96 & 50.21 & 49.73 & 49.46 & 39.69 & 47.93 \\
\hline 
\end{tabular}
\caption{Effect of two-step training. ``sent.'' denotes sentence-level parallel corpus and ``doc.'' denotes document-level parallel corpus.} \label{label:two_step}
\end{table*}

\begin{table*}[!t]
\centering
\begin{tabular}{|c||c|ccccc|c|}
\hline
Integration & MT06 & MT02 & MT03 & MT04 & MT05 & MT08 & All \\
\hline \hline
none & 48.09 & 48.63 & 47.54 & 47.79 & 48.34 & 38.31 & 45.97 \\
encoder & 48.88 & 50.30 & 49.34 & 48.81 & 49.75 & 39.55 & 47.51 \\
decoder & 49.10 & 50.31 & 49.83 & 49.35 & 49.29 & 39.07 & 47.48 \\ 
both & 49.69  & 50.96 & 50.21 & 49.73 & 49.46 & 39.69 & 47.93 \\
\hline
\end{tabular}
\caption{Effect of context integration. ``none'' means that no document-level context is integrated, ``encoder'' means that the document-level context is integrated only into the encoder, ``decoder'' means that the document-level context is integrated only into the decoder, and ``both'' means that the context is integrated into both the encoder and the decoder.} \label{table:integration}
\end{table*}

\begin{table*}[!t]
\centering
\begin{tabular}{|c||c|ccccc|c|}
\hline
Gating & MT06 & MT02 & MT03 & MT04 & MT05 & MT08 & All \\
\hline \hline
w/o & 49.33 & 50.56 & 49.74 & 49.29 & 50.11 & 39.02 & 47.55 \\
w/ & 49.69  & 50.96 & 50.21 & 49.73 & 49.46 & 39.69 & 47.93 \\
\hline
\end{tabular}
\caption{Effect of context gating.} \label{table:gating}
\end{table*}

Table \ref{label:two_step} shows the effect of the proposed two-step training strategy. The first two rows only use sentence-level parallel corpus to train the original Transformer model (see Eq. \ref{eq:sent}) and achieve BLEU scores of 39.53 and 45.97. The third row only uses the document-level parallel corpus to directly train our model (see Eq. \ref{eq:direct}) and achieves a BLEU score of 36.52. The fourth and fifth rows use the two-step strategy to take advantage of both sentence- and document-level parallel corpora and achieve BLEU scores of 40.22 and 47.93, respectively.

We find that document-level NMT achieves much worse results than sentence-level NMT (i.e., 36.52 vs. 39.53) when only small-scale document-level parallel corpora are available. Our two-step training method is capable of addressing this problem by exploiting sentence-level corpora, which leads to significant improvements across all test sets.

\subsection{Effect of Context Integration}

Table \ref{table:integration} shows the effect of integrating document-level context to the encoder and decoder (see Section \ref{sec:integration}). It is clear that integrating document-level context into the encoder (Eq. \ref{eq:ctx_enc}) brings significant improvements (i.e., 45.97 vs. 47.51). Similarly, it is also beneficial to integrate document-level context into the decoder (Eq. \ref{eq:ctx_dec}). Combining both leads to further improvements. This observation suggests that document-level context does help to improve Transformer.

\subsection{Effect of Context Gating}

As shown in Table \ref{table:gating}, we also validated the effectiveness of context gating (see Section \ref{sec:gating}). We find that replacing residual connections with context gating leads to an overall improvement of 0.38 BLEU point.

\begin{table*}[!t]
\centering
\begin{tabular}{l|p{1.5\columnwidth}}
\hline \hline
Context & {\em $\cdots$ziji ye yinwei queshao jingzheng duishou er dui \bluebf{saiche} youxie \redbf{yanjuan} shi$\cdots$}\\
\hline
Source & {\em wo rengran feichang \redbf{rezhong} yu zhexiang \bluebf{yundong}.}\\
\hline
Reference & I'm still very \redbf{fond of} the \bluebf{sport}. \\
\hline
Transformer & I am still very \redbf{enthusiastic about} this \bluebf{movement}. \\
\hline
Our work & I am still very \redbf{keen on} this \bluebf{sport}. \\
\hline \hline
\end{tabular}
\caption{An example of Chinese-English translation. In the source sentence, ``{\em yundong}'' (sport or political movement) is a multi-sense word and ``{\em rezhong}'' (fond of) is an emotional word whose meaning is dependent on its context. Our model takes advantage of the words ``{\em saiche}'' (car racing) and ``{\em yanjuan}'' (tired of) in the document-level context to translate the source words correctly.} \label{table:example}
\end{table*}

\subsection{Analysis}

We use an example to illustrate how document-level context helps translation (Table \ref{table:example}). In order to translate the source sentence, NMT has to disambiguate the multi-sense word ``{\em yundong}'', which is actually impossible without the document-level context. The exact meaning of ``{\em rezhong}'' is also highly context dependent. Fortunately, the sense of ``{\em yundong}'' can be inferred from the word ``{\em saiche}'' (car racing) in the document-level context and ``{\em rezhong}'' is the antonym of ``{\em yanjuan}'' (tired of). This example shows that our model learns to resolve word sense ambiguity and lexical cohesion problems by integrating document-level context.

\section{Related Work}

Developing document-level models for machine translation has been an important research direction, both for conventional SMT \cite{Gong:11,Hardmeier:12,Xiong:13a,Xiong:13,Garcia:14}  and NMT \cite{Jean:17,Kuang:17,Tiedemann:17,Wang:17,Maruf:18,Bawden:18,Tu:18,Voita:18}.

Most existing work on document-level NMT has focused on integrating document-level context into the RNNsearch model \cite{Bahdanau:15}. These approaches can be roughly divided into two broad categories: computing the representation of the full document-level context \cite{Jean:17,Tiedemann:17,Wang:17,Maruf:18,Voita:18} and using a cache to memorize most relevant information in the document-level context \cite{Kuang:17,Tu:18}. Our approach falls into the first category. We use multi-head attention to represent and integrate document-level context. 

\citet{Voita:18} also extended Transformer to model document-level context, but our work is different in modeling and training strategies. The experimental part is also different. While \citet{Voita:18} focus on anaphora resolution, our model is able to improve the overall translation quality by integrating document-level context.  

%The proposed two-step training strategy bears resemblance to pre-training \cite{Shen:16,Tu:18}, which is effective for training complex models on limited labeled data. We find that instead of using pre-training for model initialization, keeping the sentence-level model parameters fixed when estimating document-level parameters results in more benefits.

\section{Conclusion}
We have presented a method for exploiting document-level context inside the state-of-the-art neural translation model Transformer. Experiments on Chinese-English and French-English translation tasks show that our method is able to improve over Transformer significantly. In the future, we plan to further validate the effectiveness of our approach on more language pairs.

\section*{Acknowledgments}

Yang Liu is supported by the National Natural Science Foundation of China (No. 61432013), National Key R\&D Program of China (No. 2017YFB0202204), National Natural Science Foundation of China (No. 61761166008), Advanced Innovation Center for Language Resources  (TYR17002), and the NExT++ project supported by the National Research Foundation, Prime Ministers Office, Singapore under its IRC@Singapore Funding Initiative. This research is also supported by Sogou Inc.

\bibliography{emnlp2018_zjc}
\bibliographystyle{acl_natbib_nourl}

\end{document}